\documentclass[conference]{IEEEtran}
\IEEEoverridecommandlockouts

\usepackage[numbers,sort&compress]{natbib}
\usepackage{amsmath,amsfonts,amssymb}
\usepackage{graphicx}
\usepackage{textcomp}
\usepackage{xcolor}
\usepackage{booktabs}
\usepackage{float}
\usepackage{pifont}

\usepackage{enumitem}
\usepackage{url}
\usepackage{hyperref}
\hypersetup{
    colorlinks   = true,
    citecolor    = blue,
    urlcolor    =  blue
}
\usepackage{fontspec}
\usepackage{unicode-math}

\usepackage{tabularx}

\newcommand{\hf}[2]{\raisebox{-2.2pt}{\includegraphics[scale=0.09]{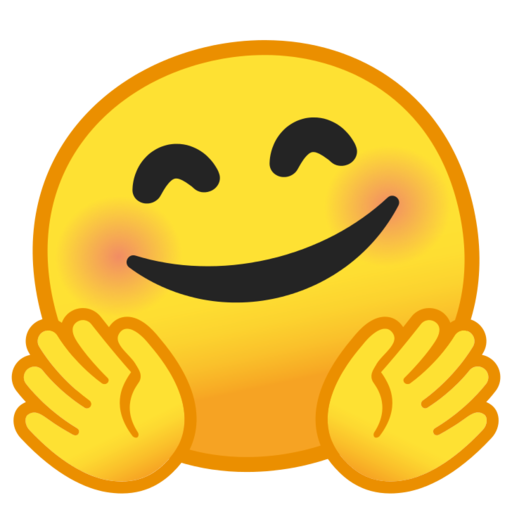}}~\href{#1}{\texttt{#2}}}

\newcommand{\gh}[2]{\raisebox{-2.2pt}{\includegraphics[scale=0.02]{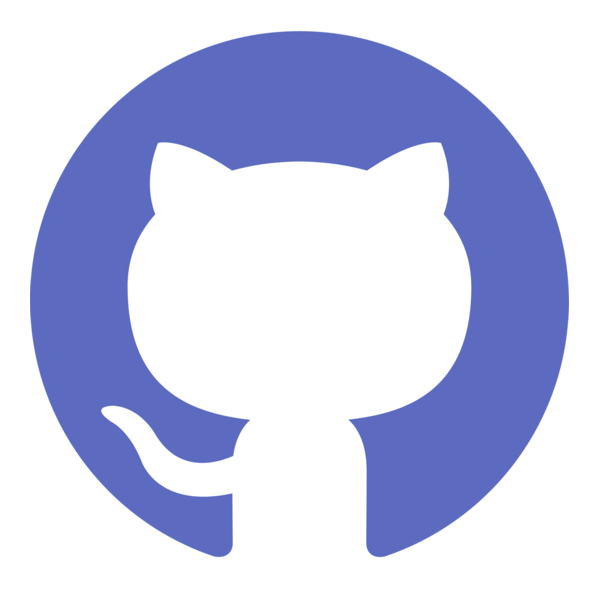}}~\href{#1}{\texttt{#2}}}

\newfontfamily\sinhalafont{NotoSansSinhala-VariableFont_wdth,wght.ttf}[Script=Sinhala]
\newcommand{\sinhala}[1]{{\sinhalafont #1}}

\newfontfamily\tamilfont{NotoSansTamil-VariableFont_wdth,wght.ttf}[Script=Tamil]
\newcommand{\tamil}[1]{{\tamilfont #1}}

\begin{document}

\title{Trilingual Topic Modeling of Sri Lankan Parliamentary Debates}

 \author{
\IEEEauthorblockN{%
Himath Dhanapala, Haren Daishika, Himandhi Kuruppu, Sithija Seneviratne, Ashini Kavindya,\\
Patalee Narasinghe, Sandeepa Weerasekara, Nisansa de Silva, Sandareka Wickramanayake 
}
\IEEEauthorblockA{Dept.\ of Computer Science \& Engineering, University of Moratuwa, Sri Lanka.\\
 \texttt{\{himathd.23, harend.23, himandhik.23, sithijas.23, ashinik.23,}\\ 
 \texttt{patalee.21, sandeepa.25, NisansaDdS, sandarekaw\}@cse.mrt.ac.lk}
 }
 }

\maketitle

\begin{abstract}
Sri Lankan parliamentary debates (Hansards) constitute a trilingual corpus of speeches in Sinhala, Tamil, and English, including code-mixed content, yet remain inaccessible to standard NLP pipelines due to layout-complex PDFs, multilingual scripts, and agglutinative morphology. We present an end-to-end framework that addresses these challenges through LLM-based text extraction followed by a multilingual embedding and density-based clustering pipeline for topic modeling. A hybrid semantic-lexical extension, BiTopic, is further explored to improve interpretability and recover speeches otherwise discarded as noise. Applied to 19,553 speeches spanning 2017–2026, the pipeline recovers 30 macro-topics achieving a cluster purity (BCP) of 0.673, whose temporal trajectories align unsupervised with major national events including the 2019 Easter Sunday attacks and the 2022 economic crisis. Traditional LDA fails on this corpus due to cross-lingual fragmentation, whereas the proposed approach successfully identifies thematic structure across all three languages without supervision.
\end{abstract}


\section{Introduction} 

The Parliament of Sri Lanka serves as the principal national forum for deliberation on governance, public policy, economic planning, and social welfare. Its official transcripts, known as Hansards, constitute one of the richest publicly available records of Sri Lankan political discourse. Published in Sinhala, Tamil, and English, these transcripts preserve legislative debate, policy argumentation, and inter-party exchange in their original linguistic form. As such, they provide a valuable resource for understanding policy priorities, ideological positioning, and the evolution of parliamentary attention across major national issues.

The systematic analysis of this corpus has both scholarly and civic value. Topic modeling of parliamentary debates can reveal how legislative attention is distributed across policy domains, track the rise and decline of issues over time, and provide a structured basis for downstream tasks such as sentiment analysis, political stance detection, and representative accountability studies. However, these benefits are difficult to realize using standard NLP pipelines without substantial adaptation.

The Sri Lankan Hansard corpus is distinguished by three interrelated challenges. First, it is inherently trilingual: speeches may be delivered in Sinhala, Tamil, English, or code-mixed combinations, sometimes within a single speaker turn. Second, Sinhala and Tamil exhibit highly agglutinative morphology~\cite{pushpananda2017improving} , causing a single lexical root to appear in many surface forms and reducing the effectiveness of bag of words representations. Third, the source documents are PDFs containing dual column layouts, mixed-script rendering, and formatting irregularities that limit conventional OCR based extraction.

These challenges motivate the development of a multilingual NLP framework specifically tailored to Sri Lankan parliamentary discourse. This study proposes an end to end pipeline combining LLM-based text extraction with multilingual embeddings and clustering-based topic modeling. In addition to a BERTopic-based pipeline, a hybrid semantic lexical BiTopic architecture is explored as an experimental extension. This work is guided by three research questions:
\begin{enumerate}[topsep=2pt, itemsep=1pt,label={\small\textbf{RQ\arabic*:}},leftmargin=3em]
    \item How can a topic modeling pipeline operate effectively over a large-scale trilingual corpus containing Sinhala, Tamil, and English, including code-mixed text, while preserving cross-lingual semantic coherence?
    \item Can macro-level thematic groupings be derived empirically from the hierarchical structure of discovered micro-topics rather than being specified arbitrarily?
    \item Does augmenting dense semantic embeddings with lexical information improve topic boundary sharpness relative to a purely embedding-based baseline?
\end{enumerate}


The main contributions of this paper are:
\begin{itemize}[topsep=2pt, itemsep=1pt]
    \item An end-to-end multilingual topic modeling pipeline for Sri Lankan Hansard data, integrating LLM-based extraction, multilingual embeddings, dimensionality reduction, density-based clustering, and topic representation.
    \item A systematic evaluation of multilingual embedding models for clustering long form, code-mixed parliamentary speeches using cross-lingual semantic similarity, semantic retrieval, and anisotropy analysis.
    \item A comparative analysis of clustering algorithms, highlighting the precision coverage trade-off and demonstrating the suitability of HDBSCAN for high-purity topic extraction.
    \item A hybrid semantic lexical BiTopic framework combining dense embeddings with lexical grounding to improve topic interpretability and recover speeches otherwise classified as noise.
    \item \hf{https://huggingface.co/datasets/himath-nimpura/sl-parliamentary-hansard-17-26}{Data} and \gh{https://github.com/HimathX/lk-hansard-topic-modeling}{code} for this work are publicly available.
\end{itemize}

\section{Related Work}
Computational analysis of parliamentary debates is widely explored using established corpora like EuroParl~\cite{koehn2005europarl}, the British Hansard~\cite{abercrombie2018sentiment}, and the U.S. Congressional Record~\cite{lauderdale2014scaling}, supporting downstream applications like machine translation, agenda tracking, and discourse analysis. Locally, Sri Lankan legislative text was recently structured into large-scale multilingual datasets~\cite{senaratna2025sri}. However, prior NLP methodologies overwhelmingly prioritize high-resource, monolingual environments. Consequently, standard pipelines struggle to process the Sri Lankan Hansard, which is uniquely trilingual and heavily code-mixed.

Traditional topic modeling has been dominated by probabilistic methods such as Latent Dirichlet Allocation (LDA)~\cite{blei2003latent}. While effective for monolingual corpora, LDA relies on bag-of-words representations, limiting its ability to capture contextual meaning and cross-lingual equivalence; these weaknesses become more severe in agglutinative languages such as Sinhala and Tamil. Recent work has shifted toward neural and embedding-based approaches. Contextualized Topic Models~\cite{bianchi2021cross} have shown improved coherence but have not been systematically evaluated on corpora combining low-resource languages and code-mixing. BERTopic~\cite{grootendorst2022bertopic} forms the methodological foundation of this work, combining contextual embedding, dimensionality reduction, density-based clustering, and class-based TF-IDF keyword extraction~\cite{alshammari2023implementation}. Its modular design suits parliamentary discourse because it does not require a pre-specified number of topics and can assign weakly related speeches to noise rather than forcing them into clusters.

The effectiveness of such models depends strongly on multilingual embeddings. Models such as multilingual-e5~\cite{wang2024multilingual}, BGE-M3~\cite{multi2024m3}, LaBSE~\cite{feng2022language}, and paraphrase-multilingual-mpnet-base-v2~\cite{song2020mpnet,reimers2019sentence} have been proposed for cross-lingual representation. For topic discovery, HDBSCAN~\cite{campello2013density,mcinnes2017hdbscan} is especially relevant because it identifies clusters of varying density while assigning outlier speeches to noise, whereas partitioning methods such as K-Means~\cite{li2012clustering} force all observations into clusters and may reduce topic purity~\cite{ikotun2023k}. Despite these developments, robust topic modeling frameworks for low-resource, code-mixed parliamentary corpora remain limited. Prior work has not addressed multilingual topic modeling for the Sri Lankan Hansard corpus at scale, nor explored hybrid semantic--lexical modeling for improving interpretability. This study addresses these gaps.

\begin{figure*}[!htb]
 \centering
 \includegraphics[width=0.9\linewidth]{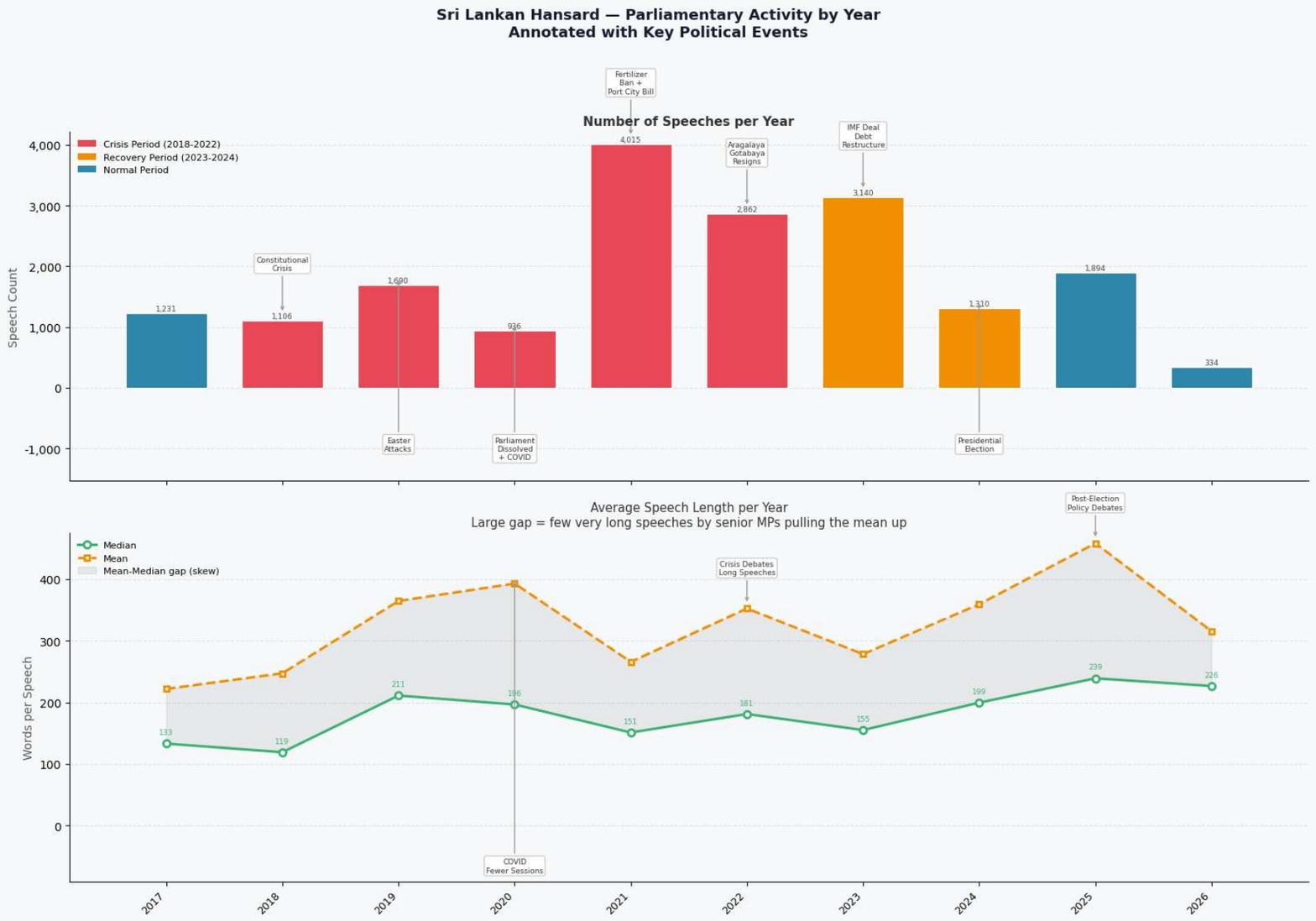}
 \caption{Parliamentary activity by year annotated with key political events.}
 \label{fig_dataset}
\end{figure*}

\begin{figure*}[!htb]
 \centering
 \includegraphics[width=0.85\linewidth]{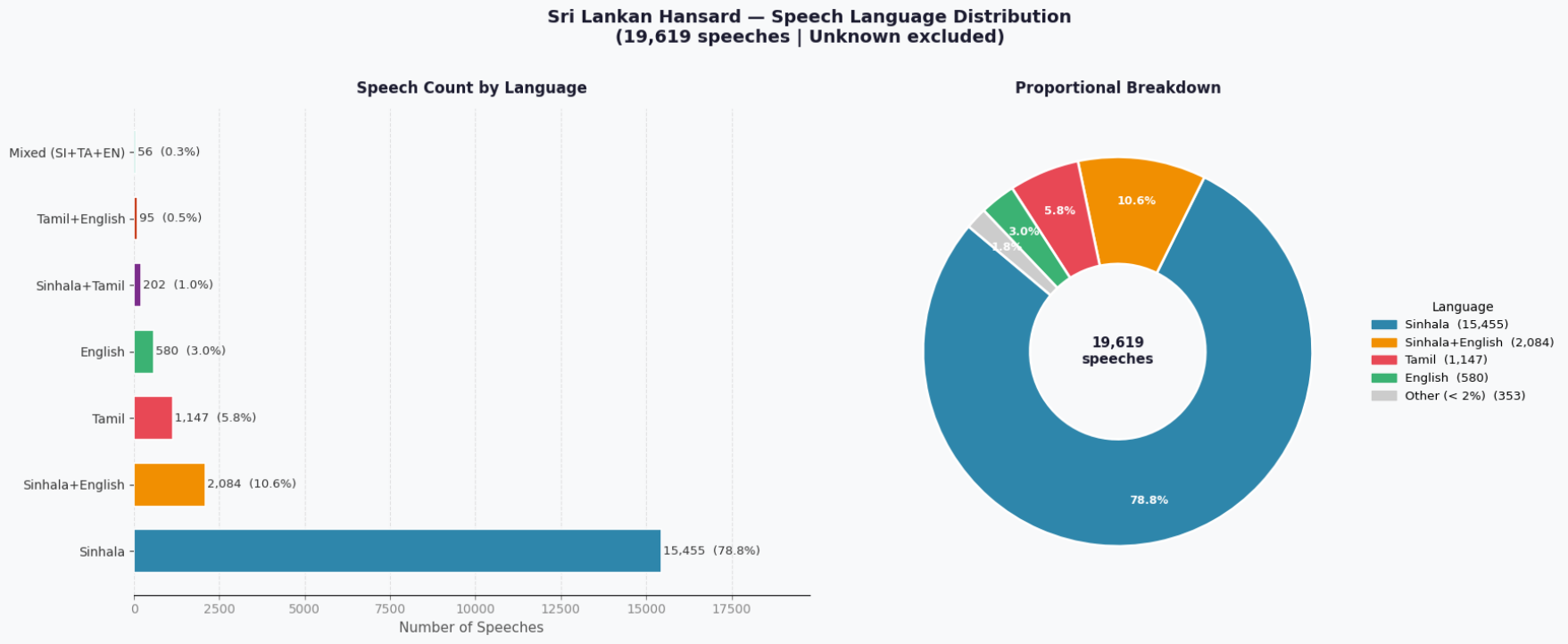}
 \caption{Speech language distribution across the Hansard corpus.}
 \label{fig_langdist}
\end{figure*}

\section{Dataset}

\subsection{Data Collection}
As Senaratna's dataset~\cite{senaratna2025sri} lacked the required multi-year coverage, we scraped Sri Lankan Hansard PDFs directly from the official parliamentary archive~\cite{Parliament2026Hansard}. A dedicated web scraping pipeline traversed year level index pages, identified linked PDFs, and downloaded them into a structured directory organized by year and sitting date. The collection spans 2017-2026 (Fig~\ref{fig_dataset}), covering both a pre-crisis baseline and the turbulent years surrounding the 2019 Easter Sunday attacks\footnote{\url{https://www.bbc.com/news/av/world-asia-48035657}}, the 2022 economic crisis\footnote{\url{https://www.bbc.com/news/world-61028138}}, the Aragalaya uprising\footnote{\url{https://www.bbc.com/news/world-asia-62108597}}, and subsequent IMF-linked restructuring\footnote{\url{https://t.co/rlpau31pw0}}. Documents not containing main debate speeches were excluded.

\subsection{LLM-Based Extraction}

Initial standard Tesseract OCR~\cite{smith2007overview} produced substantial noise on dual column layouts. Although a Sinhala fine tuned Tesseract~\cite{de2019survey} improved monolingual recognition, it failed on interspersed Tamil text. Document AI, despite being recommended for Sinhala by prior work~\cite{jayatilleke2025sidiac,jayatilleke2026sidiac}, was not used because its paid API posed a practical constraint for processing nearly 800 documents. To overcome these challenges, we adopted Gemini 3 Pro~\cite{team2023gemini}\footnote{\url{https://t.co/OcpYu64nuG}}, as part of a multimodal document extraction pipeline, which supported >1M token contexts at study initiation (February 2026)\footnote{\url{https://awesomeagents.ai/leaderboards/long-context-benchmarks-leaderboard/}}. Each PDF document was provided directly to Gemini, which performed text recognition, layout reconstruction, speaker segmentation, procedural filtering, and speaker normalization. A zero-shot prompt extracted substantive speeches, suppressed procedural noise, and mapped coherent segments to normalized speaker identifiers.

Multilingual and code mixed text was preserved without translation, transliteration, or normalization. Compared to the OCR baseline, the multimodal Gemini pipeline better handled dual column interleaving, mixed script content, and speaker attribution. To support reproducibility, the implementation and extraction pipeline, including the prompts and post processing steps, are available at \gh{https://github.com/HimathX/lk-hansard-topic-modeling/blob/docs/readme-project-team-and-audits/docs/extraction_audit/extraction_audit.md}{GitHub}.

\subsection{Ground Truth Evaluation Subset}

To enable quantitative evaluation, a manually curated subset of 300 speeches was annotated using a 16 category taxonomy grounded in the Comparative Agendas Project (CAP) \cite{baumgartner2019comparative} and extended with Sri Lanka specific categories such as Reconciliation and Disaster Management to reflect local parliamentary discourse. The final categories include Economy \& Finance, Governance \& Legal Reform, Parliamentary Affairs, Infrastructure \& Energy, Health \& Social Welfare, National Security, Agriculture \& Fisheries, Education, Foreign Affairs, Reconciliation, Labour \& Migration, Culture \& Religion, Disaster Management, Technology, Tourism, and Environmental Protection. Language identification revealed 273 Sinhala (91.0\%), 19 Tamil (6.3\%), and 8 English (2.7\%) speeches (Fig~\ref{fig_langdist}). Labels with fewer than five instances were filtered, yielding 288 speeches across 13 categories. Although the evaluation subset is Sinhala-dominant, it includes all three official parliamentary languages and therefore provides initial evidence for cross-lingual thematic clustering rather than a comprehensive validation across languages.

\begin{figure*}[h]
 \centering
 \includegraphics[width=1\linewidth]{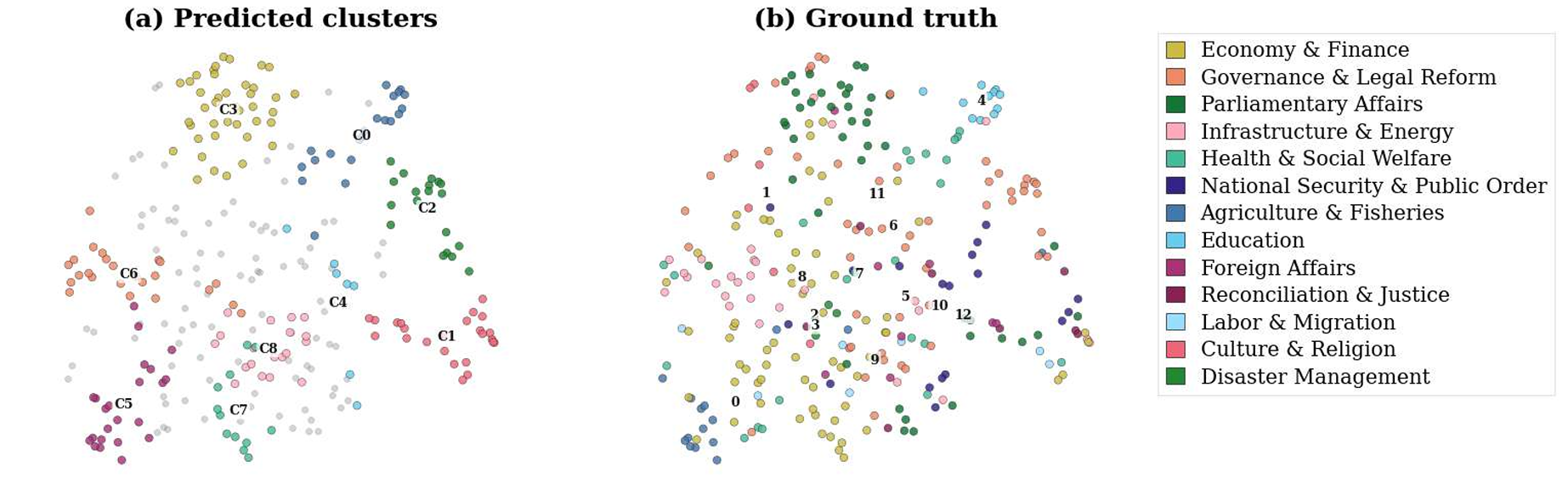}
 \caption{BiTopic framework for multilingual topic modeling.}
 \label{fig_bitopic}
\end{figure*}

\subsection{Trilingual Stopword Engineering}

A custom stopword list was developed using existing resources, statistical filtering, and parliamentary domain knowledge since available Sinhala stopword collections~\cite{wijeratne2020sinhala,lakmal2020word}, derived mainly from social media and general web corpora, did not capture Hansard-specific procedural and formulaic language.

The list was built in three stages: baseline stopwords were compiled for Sinhala, Tamil, and English; parliamentary terms such as address forms, speaker acknowledgments, and procedural motions were added; and terms occurring in more than 80\% of speeches were removed through statistical filtering. The final list comprises 1{,}045 terms.

\section{Methodology}

\subsection{Baseline: LDA}
A baseline using LDA with a custom whitespace tokenizer and the trilingual stopword list was conducted. Stemming and lemmatization were avoided because they distort non-Latin scripts. Despite these adaptations, LDA failed to produce coherent multilingual topics: its reliance on word co-occurrence limited cross-lingual semantic capture, resulting in language-specific topic fragmentation, and the morphology of Sinhala and Tamil further diluted topic distributions. This motivated the transition to embedding-based methods.

\subsection{Embedding Model Selection}
Four multilingual embedding models were evaluated on the 300-speech subset using three methods: (i)~\textbf{Cross-Lingual STS}: mean cosine similarity between speech pairs across languages within the same topic; (ii)~\textbf{Semantic Retrieval}: Precision@5 and MRR for English queries; and (iii)~\textbf{Anisotropy Analysis}: mean off diagonal cosine similarity measuring embedding space collapse.
The evaluated models include multilingual e5 large instruct, BAAI/bge-m3, LaBSE, and paraphrase-multilingual-mpnet-base-v2 (Table~\ref{table:embedding_results}) 
Due to difficulties in securing compute, LASER3~\cite{heffernan-etal-2022-bitext} had to be excluded from the study despite previous work~\cite{fernando-etal-2025-improving,ranathunga-etal-2024-quality} showing that LASER3 outperforms LaBSE for Sinhala.

\begin{table}[h]
\centering
\renewcommand{\arraystretch}{1.15}
\caption{Embedding model benchmark results on 300 Hansard speeches.}
\label{table:embedding_results}
\resizebox{\columnwidth}{!}{%
\begin{tabular}{|l|c|c|c|c|l|}
\hline
\textbf{Model} & \textbf{STS $\uparrow$} & \textbf{MRR $\uparrow$} & \textbf{P@5 $\uparrow$} & \textbf{Aniso. $\downarrow$} & \textbf{Verdict} \\ \hline
multilingual-e5-large & 0.827 & 0.810 & 0.56 & 0.909 & \ding{55} Collapsed \\ \hline
\textbf{BAAI/bge-m3} & \textbf{0.523} & \textbf{0.516} & \textbf{0.68} & \textbf{0.552} & \textbf{\ding{51} Selected} \\ \hline
LaBSE & 0.487 & 0.324 & 0.72 & 0.492 & \ding{51} Healthy \\ \hline
mpnet-base-v2 & 0.424 & 0.536 & 0.76 & 0.496 & \ding{51} Healthy \\ \hline
\end{tabular}%
}
\end{table}

\begin{table}[h]
\centering
\renewcommand{\arraystretch}{1.15}
\caption{Clustering based embedding evaluation on 300 Hansard speeches.}
\label{table:embedding_clustering}
\resizebox{\columnwidth}{!}{%
\begin{tabular}{|l|c|c|c|c|c|c|}
\hline
\textbf{Model} & \textbf{Dim} & \textbf{Topics} & \textbf{Outlier \%} & \textbf{Silh.} & \textbf{DB $\downarrow$} & \textbf{$c_v$} \\ \hline
LaBSE & 768 & 3 & 0.28 & 0.90 & 3.10 & 0.46 \\ \hline
mpnet-base-v2 & 768 & 16 & 18.97 & 0.79 & 3.09 & 0.25 \\ \hline
\textbf{BAAI/bge-m3} & \textbf{1024} & \textbf{89} & \textbf{27.57} & \textbf{0.81} & \textbf{2.48} & \textbf{0.38} \\ \hline
multilingual-e5-large & 1024 & 70 & 38.63 & 0.71 & 3.03 & 0.04 \\ \hline
\end{tabular}%
}
\end{table}

Despite its higher STS score, multilingual e5's anisotropy of 0.909 collapses embeddings into a near uniform cone, eliminating the density contrasts UMAP and HDBSCAN require. Furthermore, its 512-token limit would truncate approximately 35-40\% of parliamentary speeches. BGE-M3's anisotropy score of 0.552, combined with its 8{,}192-token context window (reducing truncation to under 3\%), made it the only viable choice for density-based clustering of long-form text, preserving discourse-level semantics critical for ministerial addresses (Table~\ref{table:embedding_clustering}). 

\subsection{Dimensionality Reduction}
UMAP~\cite{mcinnes2018umap} was applied to reduce 1024-dimensional embeddings to 5 dimensions ($n\_neighbors{=}15$, $n\_components{=}5$, $min\_dist{=}0.0$, cosine distance). UMAP preserves local neighbourhood relationships while reshaping the space to produce clearer density variations, enabling HDBSCAN to identify coherent clusters. The low $min\_dist$ value encourages tighter grouping of semantically similar points. A separate 2D projection ($min\_dist{=}0.1$) was generated for visualization.

\subsection{Clustering Algorithm Comparison}
Four clustering algorithms were evaluated on UMAP-reduced embeddings using the 288 speech subset (Table~\ref{table:clustering_results}). For density-based methods, noise points (label $-1$) were treated as singleton clusters in B-Cubed computation to ensure fair comparison with partition-based methods.

\begin{table}[h]
\centering
\renewcommand{\arraystretch}{1.15}
\caption{Clustering algorithm benchmark (288 speeches, 13 true topics).}
\label{table:clustering_results}
\resizebox{\columnwidth}{!}{%
\begin{tabular}{|l|c|c|c|c|c|c|c|}
\hline
\textbf{Algorithm} & \textbf{K} & \textbf{Noise \%} & \textbf{BCP} & \textbf{BCF1} & \textbf{ARI} & \textbf{NMI} & \textbf{Silh.} \\ \hline
K-Means & 13 & 0.0 & 0.349 & 0.324 & 0.184 & 0.379 & 0.388 \\ \hline
Agglom. (Ward) & 13 & 0.0 & 0.355 & 0.340 & 0.191 & 0.398 & 0.377 \\ \hline
\textbf{HDBSCAN} & \textbf{11} & \textbf{41.7} & \textbf{0.673} & \textbf{0.313} & \textbf{0.104} & \textbf{0.340} & \textbf{0.532} \\ \hline
OPTICS & 19 & 36.1 & 0.670 & 0.250 & 0.073 & 0.351 & 0.497 \\ \hline
\end{tabular}%
}
\end{table}

The results reveal a clear trade-off between cluster purity and coverage. Partition-based methods assign all speeches into clusters, achieving higher B-Cubed F1 through full coverage. HDBSCAN, in contrast, achieved higher cluster purity (BCP~=~0.673) and geometric separation (Silhouette~=~0.532), but at the cost of excluding 41.7\% of speeches as noise. Thus, HDBSCAN is preferable when high-purity topic discovery is more important than full coverage. Partitioning methods provide broader coverage, whereas HDBSCAN yields cleaner retained clusters by filtering procedurally weak or low-signal speeches.

\subsection{Final BERTopic Pipeline}
The final pipeline integrates BGE-M3, UMAP, and HDBSCAN within BERTopic. A \texttt{CountVectorizer}~\cite{pedregosa2011scikit}, with the trilingual stopword list retains meaningful features while suppressing boilerplate vocabulary. Topic representations were generated using class-based TF-IDF together with KeyBERT-inspired and Maximal Marginal Relevance techniques for keyword diversity and interpretability. Applied to the full corpus of 19{,}553 speeches, the pipeline produced 336 micro-topics, assigning 6{,}921 speeches (35.4\%) to noise and retaining 12{,}632 in substantive clusters.

\subsection{Macro Topic Aggregation}
The 336 micro-topics were too numerous for higher-level interpretation. Micro-topics were first ranked by size and filtered using a Pareto-style threshold to retain the dominant clusters representing most substantive speeches. Centroids of the retained micro-topics were then grouped using average-linkage hierarchical clustering with cosine distance. Rather than fixing $K$ \textit{a priori}, the largest merge-distance gap within an interpretable range $[K_{min}, K_{max}]$ was selected, yielding a data-driven cut at $0.002$ and $K_{macro}=30$. Complete implementation details are available in the \href{https://github.com/HimathX/lk-hansard-topic-modeling/tree/main/docs/macro_topic_aggregation}{project repository}.

\subsection{Experimental BiTopic Framework}
For this study BiTopic is presented as an exploratory extension rather than the main contribution. It combines semantic and lexical similarity before topic formation. For each speech, a dense BGE-M3 embedding and a sparse CountVectorizer vector are constructed. Pairwise cosine similarities are computed independently in both spaces, normalized to a common scale, and combined through weighted fusion. Early fusion combining similarities before clustering rather than merging labels post hocensures HDBSCAN operates on a unified geometric field jointly encoding both modalities. The fused similarity uses an asymmetric scheme with semantic similarity given dominant weight. This reflects a meaning-first strategy while preserving enough lexical structure to avoid conflating semantically adjacent but institutionally distinct speeches.

\section{Results}

\subsection{Macro-Topic Discovery}
\begin{figure*}[h]
 \centering
 \includegraphics[width=\linewidth]{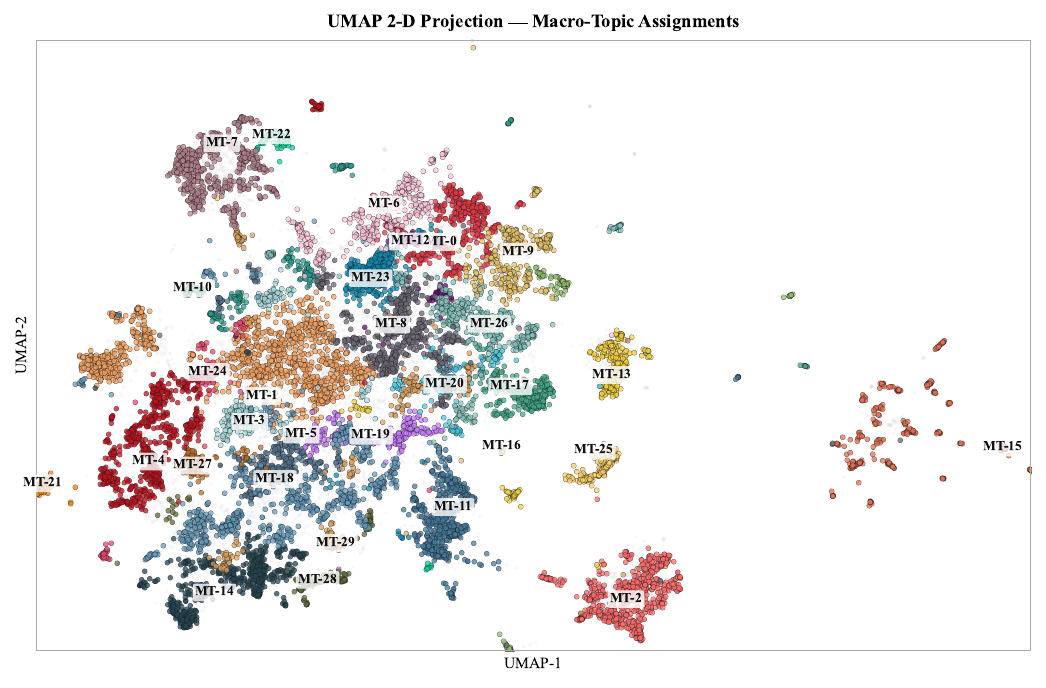}
 \caption{Macro topic assignments in 2D UMAP space.}
 \label{fig_macroscatter}
\end{figure*}

\begin{figure*}
    \centering
    \includegraphics[width=1.0\linewidth]{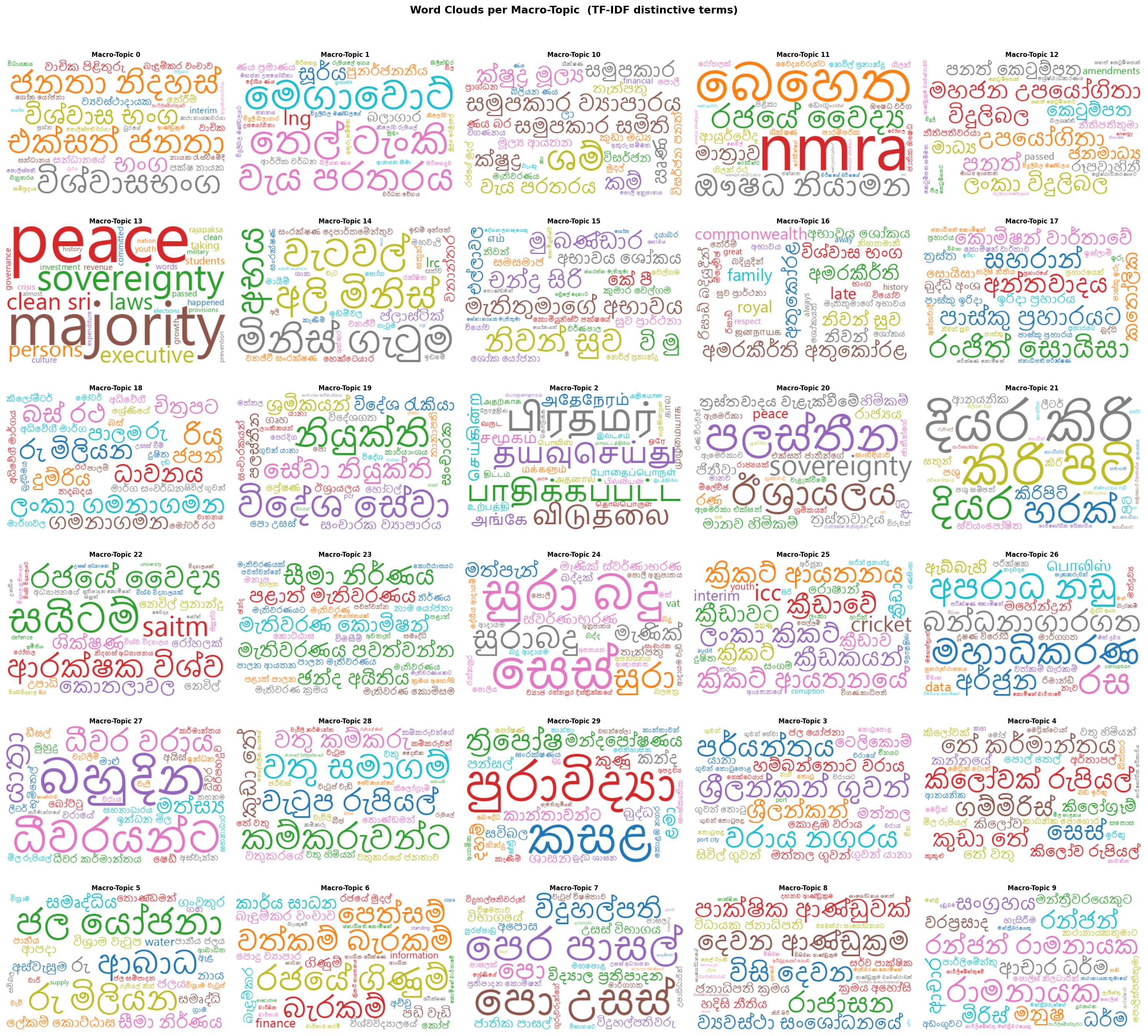}
    \caption{Macro Topic Wordclouds. (Colors are decorative only and carry no semantic meaning.)}
    \label{fig:MacroCloud}
\end{figure*}

The taxonomy spans expected high-salience domains alongside politically specific niches that would be subsumed under coarser models (Table~\ref{table:discovered_topics_new}). Economy-related topics dominate the corpus (Fig~\ref{fig_macroscatter}), reflecting Sri Lanka's sustained fiscal and energy crises. Institutional topics capture repeated debates around executive power, electoral reform, and public oversight (Fig~\ref{fig:MacroCloud}). 

Crucially, MT-17 isolates Easter Sunday attack accountability discourse via highly specific terms like \sinhala{සහරාන්} (Saharan) and \textit{commission}, while MT-22 captures the SAITM medical education controversy entirely. Equal sized topics would signal artificial balance the observed volume variation (from 89 speeches in MT-12 to 1{,}395 speeches in MT-1) reflects genuine agenda asymmetry.

\begin{table}[h]
\centering
\caption{Discovered macro-topics on Hansard speeches.}
\label{table:discovered_topics_new}
\setlength{\tabcolsep}{10pt}
\renewcommand{\arraystretch}{1.8}
\resizebox{\columnwidth}{!}{
\large  
\begin{tabular}{|c|c|l|l|}
\hline
\textbf{ID} & \textbf{N} & \textbf{Interpreted Label} & \textbf{Key Terms} \\ \hline
MT-0 & 509 & Legislative Mechanics \& Standing Orders & \sinhala{ප්‍රශ්න, ආණ්ඩුක්‍රම} \\ \hline
MT-1 & 1395 & Energy Security \& Macroeconomic Crisis & \sinhala{රුපියල්, ණය, ආර්ථික} \\ \hline
MT-2 & 738 & National Security, Ethno-Religious Tensions & \tamil{பயங்கரவாதத், தேர்தல், பௌத்த} \\ \hline
MT-3 & 340 & Mega-Infrastructure, Ports \& Aviation Development & \sinhala{වරාය නගරය, හම්බන්තොට} \\ \hline
MT-4 & 736 & Agricultural Subsidies, Food Security \& Price Controls & \sinhala{තේ, පොහොර, කාබනික} \\ \hline
MT-5 & 852 & Public Welfare, Housing \& Water Sanitation & \sinhala{නිවාස, ජල, සමෘද්ධි} \\ \hline
MT-6 & 424 & State Enterprise Accountability \& COPE Oversight & \sinhala{රජයේ ගිණුම්,} COPE \\ \hline
MT-7 & 724 & Educational Administration \& Human Capital Development & \sinhala{පාසල්, ගුරු, විශ්ව විද්‍යාල} \\ \hline
MT-8 & 568 & Constitutional Reform \& Executive Powers & \sinhala{ව්‍යවස්ථා, සංශෝධනය} \\ \hline
MT-9 & 360 & Parliamentary Privilege \& Member Conduct & \sinhala{රාමනායක, වරප්‍රසාද} \\ \hline
MT-10 & 227 & Rural Debt Crisis, Microfinance \& Cooperatives & \sinhala{සමුපකාර, ක්ෂුද්‍ර මූල්‍ය} \\ \hline
MT-11 & 685 & Public Healthcare System \& Pharmaceutical Regulation & nmra, \sinhala{ඖෂධ, කොවිඩ්} \\ \hline
MT-12 & 89 & Public Utilities \& Media Broadcasting Regulation & \sinhala{විදුලිබල, ජනමාධ්‍ය} \\ \hline
MT-13 & 317 & Foreign Policy, Peacebuilding \& Transitional Justice & peace, sovereignty \\ \hline
MT-14 & 653 & Environmental Conservation \& Land Rights Disputes & \sinhala{වනජීවී, අලි-මිනිස්} \\ \hline
MT-15 & 481 & State Tributes \& Parliamentary Condolences & \sinhala{නිවන් සුව, අභාවය} \\ \hline
MT-16 & 184 & Partisan Politics \& Adjournment Resolutions & Commonwealth, Elizabeth \\ \hline
MT-17 & 439 & Easter Sunday Attacks \& State Intelligence & \sinhala{පාස්කු, සහරාන්} \\ \hline
MT-18 & 370 & Road Infrastructure \& Public Transport Networks & \sinhala{දුම්රිය, අධිවේගී} \\ \hline
MT-19 & 234 & Migrant Labour, Tourism \& Foreign Exchange & \sinhala{විදේශ රැකියා} \\ \hline
MT-20 & 189 & International Relations \& Human Rights Frameworks & \sinhala{මානව හිමිකම්, ජිනීවා} \\ \hline
MT-21 & 90 & Livestock Development \& Dairy Production & \sinhala{කිරි පිටි, දියර කිරි} \\ \hline
MT-22 & 114 & Private Higher Education (SAITM) \& University Protests & SAITM, \sinhala{කොතලාවල} \\ \hline
MT-23 & 389 & Local Government Governance \& Electoral Reforms & \sinhala{මැතිවරණ, ඡන්ද} \\ \hline
MT-24 & 148 & State Revenue, Excise Policy \& Gem Industry & VAT, \sinhala{සුරා බදු} \\ \hline
MT-25 & 240 & National Sports Administration \& Cricket Board Crises & ICC, \sinhala{ක්‍රිකට්} \\ \hline
MT-26 & 563 & Judicial System, Anti-Corruption \& Criminal Law & \sinhala{අධිකරණ, දූෂණ} \\ \hline
MT-27 & 133 & Maritime Boundaries \& Fisheries Livelihoods & \sinhala{ධීවර, මත්ස්‍ය} \\ \hline
MT-28 & 155 & Plantation Economy \& Estate Worker Grievances & \sinhala{වතු, වැටුප} \\ \hline
MT-29 & 286 & Women \& Child Affairs, Heritage \& Local Governance & \sinhala{පුරාවිද්‍යා, කාන්තාවන්} \\ \hline
\end{tabular}%
}
\end{table}

\subsection{Temporal and Cross Lingual Patterns}
\begin{figure*}[t]
    \centering
    \includegraphics[width=1.0\textwidth]{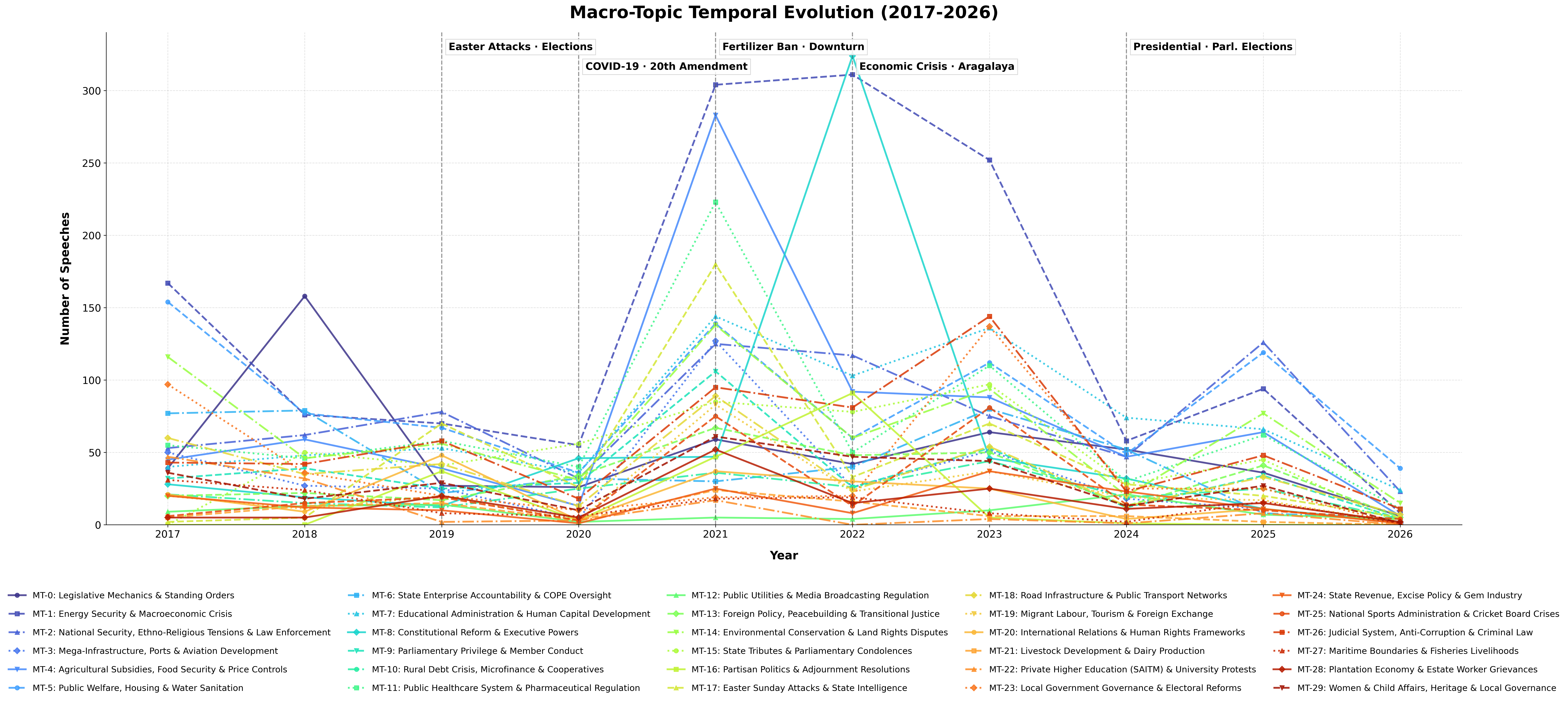}
    \caption{Temporal Evolution of Macro Topics}
    \label{fig:mactoTemp}
\end{figure*}

Temporal trajectories of the discovered topics align closely with major national events (Fig~\ref{fig:mactoTemp}). Economy linked topics show clear surges during the 2022 economic crisis and Aragalaya period, while the Easter Attacks topic rises sharply around 2019 and the subsequent period of parliamentary scrutiny. Healthcare related topics show distinct prominence during the COVID-19 period. Crucially, no event labels were provided at any stage of training or clustering; the temporal alignment emerges purely from discourse structure, confirming that the model tracks substantive shifts in legislative attention rather than surface vocabulary repetition.

The language distribution across macro-topics further shows that most topics contain speeches from more than one language. While a small number of topics are more strongly associated with Tamil-medium or English medium debate, the majority of topics reflect clustering by thematic content rather than by language identity.

\subsection{Clustering Quality and BiTopic Results}
Quantitative evaluation confirms that the embedding-based pipeline outperforms traditional bag-of-words baselines. HDBSCAN achieved the strongest cluster purity and geometric separation among the methods tested. The embedding benchmark similarly showed that BGE-M3 provided a more suitable representation space for clustering than the alternatives (Fig~\ref{fig_bitopic}).

For the BiTopic framework, an ablation over $\alpha \in [0.60, 1.00]$ showed that $\alpha = 0.85$ ($\beta = 0.15$) corresponds to an inflection in the noise coverage curve: consolidating 336 semantic micro-topics into 277 sharper clusters, albeit at a measurable coverage cost (noise increased from 35.4\% to 46.0\%). The semantic BERTopic pipeline remained the stronger main model, while BiTopic is best viewed as an exploratory direction for future refinement. A more comprehensive visualization of the results is available through our dashboard {\url{https://hansards.vercel.app/}}.

\section{Discussion}
The thematic coherence of 30 unsupervised macro-topics, whose temporal trajectories align with independently verifiable national events, demonstrates that Sri Lankan parliamentary attention is sufficiently structured for embedding-based modeling to recover politically meaningful discourse boundaries without any supervision signal.
The cross-lingual composition of most macro-topics carries a broader implication for low-resource NLP: when semantic embeddings are strong enough to overcome language-surface variation, thematic content dominates cluster formation over language identity. This validates the use of multilingual embeddings as a unifying representation layer for code-mixed, multi-script corpora where translation or language separation preprocessing would destroy analytically significant content.
The macro-topic aggregation result further suggests that parliamentary discourse organizes hierarchically micro-topics capture procedural and session-specific variation, while macro-topics recover stable policy domains. The BiTopic coverage purity tradeoff indicates that lexical grounding is most valuable for institutionally distinct but semantically adjacent debate domains.

\section{Limitations and Future Work}
This study has certain limitations. First, the supervised evaluation relies on a small ground truth subset of 300 speeches heavily skewed towards Sinhala. Second, topics with fewer than five speeches were filtered prior to evaluation, reflecting the genuine imbalance of parliamentary time allocation. Third, HDBSCAN's noise rejection rate of 41.7\% on the benchmark means a substantial fraction of parliamentary time is excluded. While appropriate for sentiment analysis, this represents a coverage loss for research requiring complete corpus analysis.

Future work will focus on two directions: (1)~Aspect-Based Sentiment Analysis (ABSA) within coherent topic clusters to analyze how political actors respond to specific policy issues, and (2)~Temporal Sentiment Tracking across parliamentary sessions to understand the dynamic nature of political discourse.

\section{Conclusion}
This study presented a multilingual topic modeling framework for Sri Lankan parliamentary debates, combining LLM based text extraction, multilingual embeddings, density-based clustering, and hierarchical aggregation to recover thematic structure from a noisy trilingual corpus. The pipeline overcomes the main limitations of bag-of-words approaches regarding cross lingual semantic fragmentation, agglutinative morphology, and long-form parliamentary text. The combination of BGE-M3, UMAP, and HDBSCAN provides a robust basis for extracting semantically coherent and politically interpretable topics. The 30 macro-topics whose temporal patterns align with major political and economic events confirm the substantive validity of the approach. The exploratory BiTopic extension showed that lexical grounding can improve topic boundary sharpness, though with a trade-off in coverage. Overall, this work establishes a strong foundation for multilingual parliamentary NLP in Sri Lanka and provides a structured basis for downstream tasks such as sentiment analysis, stance detection, and longitudinal analysis of legislative attention.

{\footnotesize
\bibliographystyle{IEEEtranN}
\bibliography{references}
}

\end{document}